\documentclass[10pt,letterpaper,twocolumn]{article}
\usepackage[latin9]{inputenc}
\usepackage{array}
\usepackage{multirow}
\usepackage{amsmath}
\usepackage{amssymb}
\usepackage{graphicx}
\usepackage{rotating}
\usepackage[unicode=true,
 bookmarks=false,
 breaklinks=true,pdfborder={0 0 1},backref=section,colorlinks=false]
 {hyperref}
\hypersetup{
 pagebackref=true,letterpaper=true,colorlinks}

\makeatletter

\pdfpageheight\paperheight
\pdfpagewidth\paperwidth

\providecommand{\tabularnewline}{\\}

\usepackage{iccv}
\usepackage{times}
\usepackage{epsfig}
\usepackage{graphicx}

\iccvfinalcopy 

\def\iccvPaperID{9107} 

\ificcvfinal\fi

\makeatother

\begin{document}

\title{G-DetKD: Towards General Distillation Framework for Object Detectors
via Contrastive and Semantic-guided Feature Imitation}
\author{Lewei Yao$^{1}$\thanks{Equal contribution} \hspace{4mm}Renjie Pi$^{1*}$\hspace{4mm}Hang
Xu$^{2}$\thanks{Corresponding author:\textit{ }xbjxh@live.com}\hspace{4mm}Wei
Zhang$^{2}$ \hspace{4mm}Zhenguo Li$^{2}$\hspace{4mm}Tong Zhang$^{1}$\\
$^{1}$Hong Kong University of Science and Technology\hspace{4mm}$^{2}$Huawei
Noah's Ark Lab}

\maketitle
\ificcvfinal\thispagestyle{empty}\fi


In this paper, we investigate the knowledge distillation (KD) strategy
for object detection and propose an effective framework applicable
to both homogeneous and heterogeneous student-teacher pairs. The conventional
feature imitation paradigm introduces imitation masks to focus on
informative foreground areas while excluding the background noises.
However, we find that those methods fail to fully utilize the semantic
information in all feature pyramid levels, which leads to inefficiency
for knowledge distillation between FPN-based detectors. To this end,
we propose a novel semantic-guided feature imitation technique, which
automatically performs soft matching between feature pairs across
all pyramid levels to provide the optimal guidance to the student.
To push the envelop even further, we introduce contrastive distillation
to effectively capture the information encoded in the relationship
between different feature regions. Finally, we propose a generalized
detection KD pipeline, which is capable of distilling both homogeneous
and heterogeneous detector pairs. Our method consistently outperforms
the existing detection KD techniques, and works when (1) components
in the framework are used separately and in conjunction; (2) for both
homogeneous and heterogenous student-teacher pairs and (3) on multiple
detection benchmarks. With a powerful X101-FasterRCNN-Instaboost detector
as the teacher, R50-FasterRCNN reaches 44.0\% AP, R50-RetinaNet reaches
43.3\% AP and R50-FCOS reaches 43.1\% AP on COCO dataset.


\section{Introduction}

\begin{figure}
\begin{centering}
\includegraphics[scale=0.55]{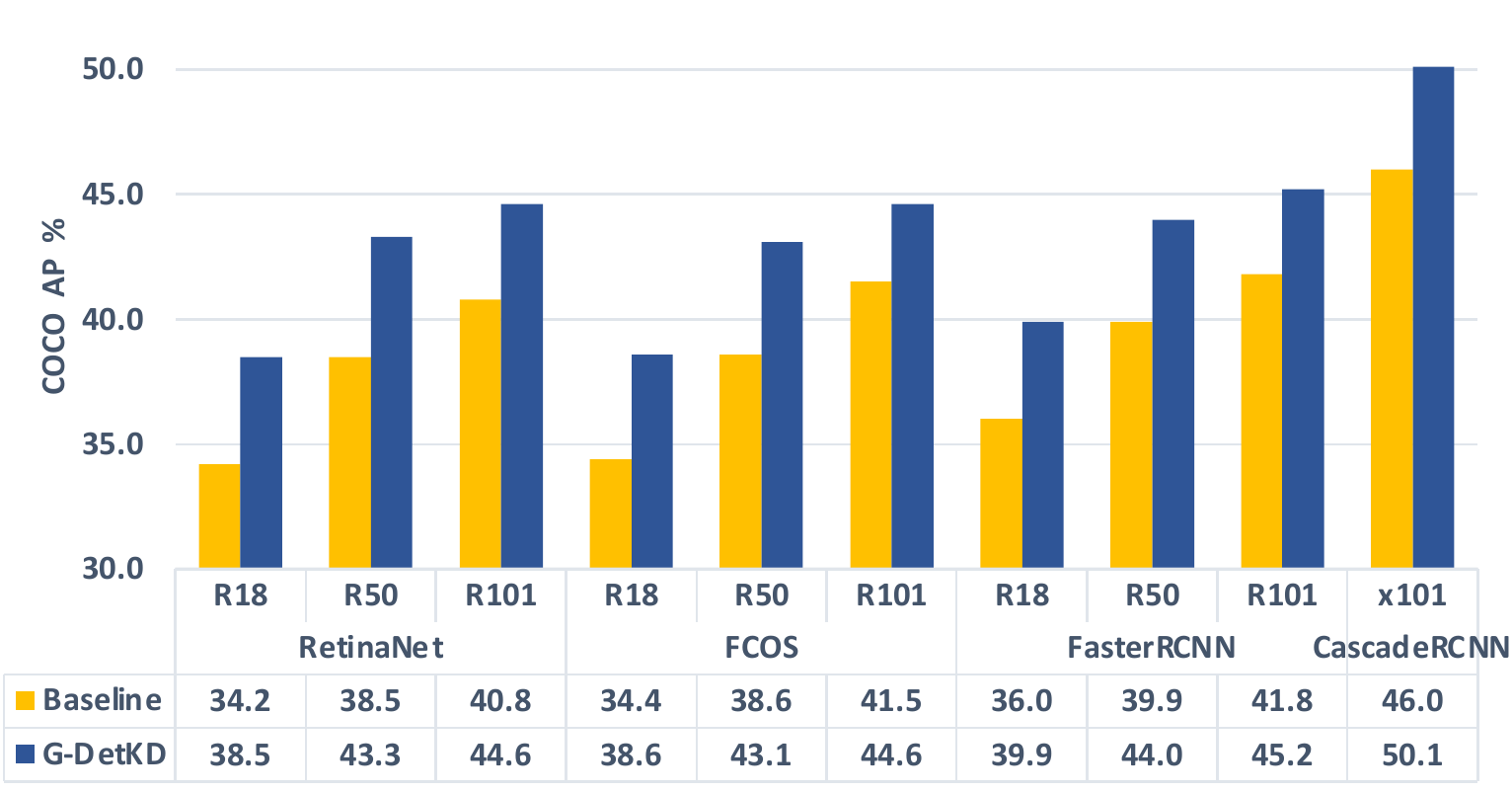}
\par\end{centering}
\caption{\label{fig:performances}Our G-DetKD consistently improves the performance
of detectors belonging to various categories by around 4\% in AP.}
\end{figure}

Knowledge distillation (KD) is a training strategy aiming at transferring
the learnt knowledge from a powerful, cumbersome model (teacher) to
a more compact model (student). The seminal work \cite{hinton2015distilling}
introduced the idea of KD and this technique has been proven to be
effective on classification tasks by many subsequent works \cite{zagoruyko2016paying,furlanello2018born,tarvainen2017mean}.
However, integrating the idea of KD into detection is nontrivial.
The conventional paradigm of classification KD can not be directly
applied to detection task since simply minimizing the KL divergence
between the classification outputs fails to extract the spatial information
from the teacher and only brings limited performance gain \cite{wang2019distilling}
to the student. In this work, we aim at developing a general KD framework
which can efficiently extract the spatial information and is applicable
to both homogeneous and heterogeneous student-teacher pairs.

It is acknowledged that feature map imitation with foreground attention
mechanisms helps students learn better \cite{chen2017learning,Li_2017_CVPR,wang2019distilling,sun2020distilling}.
Previous works propose different imitation masks to focus on the informative
foreground regions while excluding the background noises. However,
mask-based methods were first developed for outdated detectors, i.e.,
vanilla Faster-RCNN without FPN \cite{Ren2015}, which fail to extend
to modern detectors equipped with FPN. Specifically, those methods
perform direct one-to-one matching between pyramid levels of the student-teacher
pair, which leads to two issues: (1) indiscriminately applying the
same mask on all levels can introduce noise from unresponsive feature
levels; (2) mask-based methods are not extendible to heterogeneous
detector pairs since their feature levels may not be strictly aligned,
e.g., FasterRCNN constructs the feature pyramid from $P2$ to $P6$,
while RetinaNet uses $P3$ to $P7$.

To address the above issues, we propose a simple yet effective Semantic-Guided
Feature Imitation (\textbf{SGFI}) approach based on object proposals.
By analyzing the response patterns of feature pyramids as shown in
Figure \ref{fig:featmap}, we find that the best matching features
for imitation may be from different pyramid levels for the student-teacher
pair. In addition, the object regions on nearby feature levels are
activated in a similar pattern, which implies that student features
from other levels carrying similar semantics can also benefit from
imitation. Thus, to fully exploit the potential of teacher's feature
pyramid, our proposed method automatically performs soft matching
between the features based on their semantic relationship to provide
the student with the optimal guidance. The proposed SGFI outperforms
other feature imitation approaches by a large margin.

Feature imitation techniques mimic the features corresponding to the
same region, whereas the relation between representations of different
regions also encodes informative knowledge to help the student's learning.
Thus, we further propose Contrastive Knowledge Distillation (\textbf{CKD})
for detection inspired by the idea of contrastive learning. Specifically,
we make use of the proposal region's representations produced by the
student-teacher pair to construct positive and negative contrastive
pairs, then an InfoNCE loss is adopted to minimize the distance between
the positive pairs while pushing away the negative ones. We show that
the standalone contrastive approach demonstrates outstanding effectiveness
and can further boost the detector's performance when applied with
our SGFI.

In some scenarios, only detectors with certain architectures can be
deployed due to hardware constraints, while the most powerful teachers
belong to different categories. In this case, knowledge distillation
between heterogeneous detector pairs is promising. However, previous
works only consider KD between homogeneous detectors pairs due to
their lack of extensibility. Therefore, we are motivated to extend
the two approaches mentioned above into a general detection KD framework
called \textbf{G-DetKD} which is applicable to both homogeneous and
heterogeneous student-teacher detector pairs.

Extensive experiments demonstrate the effectiveness of our proposed
G-DetKD framework. On COCO dataset \cite{Lin2014}, without bells
and whistles, G-DetKD easily achieves state-of-the-art performance
among detection KD methods. As \textbf{no modification is applied
to the student detectors}, \textbf{the performance gain is totally
free}. Using the same two-stage detector as the teacher, the performance
gains for homogeneous students surpass other SOTA detection KD methods:
R50-FasterRCNN reaches 44.0\% AP; while the effect on heterogeneous
students is also significant: R50-Retina reaches 43.3 \% AP and R50-FCOS
reaches 43.1\% AP. Our method generalizes surprisingly well for large
detectors like CascadeRCNN with ResNeXt101-DCN as the backbone: boosting
its AP from 46.0\% to 50.5\%. In addition, the generalization ability
of our method is validated on multiple mainstream detection benchmarks,
e.g., Pascal VOC \cite{Everingham10} and BDD \cite{Yu2018}.

In summary, the contributions of this paper are threefold:
\begin{itemize}
\item We propose a novel semantic-guided feature imitation approach (SGFI)
with a semantic-aware soft-matching mechanism.
\item We propose contrastive knowledge distillation (CKD) to capture the
information encoded in the relationship between teacher's different
feature regions.
\item We make the first attempt to construct a general KD framework (G-DetKD)
capable of distilling knowledge for both homogeneous and heterogeneous
detectors pairs. Comprehensive experiments are conducted to show the
significant performance boosts brought by our approach.
\end{itemize}

\section{Related Works}

\textbf{Object Detection}. Object detection is one of the fundamental
problems in computer vision. State-of-the-art detection networks can
be categorized into one-stage, two-stage and anchor-free detectors.
One-stage detectors such as \cite{redmon2017yolo9000,Liu2016,redmon2018yolov3}
perform object classification and bounding box regression directly
on feature maps. On the other hand, two-stage detectors such as \cite{Ren2015,Lin2017a}
adopt a ``coarse-to-fine'' approach which uses a region proposal
network (RPN) to separate foreground boxes from background and a RCNN
head to refine the regression results and perform the final classification.
\cite{tian2019fcos,duan2019centernet,yang2019reppoints} propose to
directly predict location of objects rather than based on anchor priors,
which opens a new era for object detection. Recent works also perform
NAS on detection tasks, which searches for novel detectors automatically
without human intervention \cite{jiang2020sp,xu2019auto,Ghiasi_2019_CVPR,yao2021jointdetnas,yao2019sm}.

\textbf{Knowledge Distillation.} KD was first proposed in \cite{hinton2015distilling}.
Its effectiveness has been explored by many subsequent works \cite{zagoruyko2016paying,furlanello2018born,tarvainen2017mean,romero2014fitnets}.
For object detection, \cite{chen2017learning} first proposes imitating
multiple components in detection pipeline, e.g., backbone features,
RPN and RCNN. Recent works demonstrate that foreground instances are
more important in feature imitation. Various methods are proposed
to help student model focus on foreground information, including multiple
mask generating approaches \cite{tian2019contrastive,sun2020distilling}
and RoI extraction \cite{Li_2017_CVPR}. However, these methods are
designed for detectors without FPN, which fail to extend to FPN-based
detectors with multiple feature levels. This motivates us to design
a new KD framework that is able to fully exploit teacher's feature
pyramid.

\textbf{Contrastive Learning}. Contrastive learning is a popular approach
for self-supervised tasks \cite{oord2018representation,he2020momentum,chen2020simple}.
The goal is to bring closer the representations of similar inputs
and push away those of dissimilar ones, which naturally takes into
account the relationship between contrastive pairs. Inspired by the
recent work \cite{tian2019contrastive}, which proposes a distillation
method using a contrastive approach for classification, we propose
to integrate the information encoded in the relationship between different
object regions by introducing contrastive knowledge distillation for
object detection.

\section{Preliminary\label{sec:Preliminary}}

We start by briefly reviewing previous works on KD for object detection.
\cite{chen2017learning} conducted knowledge distillation on detector's
classification and localization predictions, which are formulated
as: $L_{cls}=-\frac{1}{N}\sum^{N}\mathbf{P}_{t}\log\mathbf{P}_{s}$,
$L_{reg}=-\frac{1}{N}\sum^{N}\mid reg_{t}^{i}-reg_{s}^{i}\mid$, where
$P_{s}$, $P_{t}$ and $reg_{s}$, $reg_{t}$ are the class scores
and localization outputs of the student-teacher pair, respectively.
However, simply distilling from the prediction outputs neglects the
teacher's spatial information.

More recent works mainly focus on distilling knowledge from the feature
maps, which encode spatial information that is crucial for detection.
These methods propose different imitation masks $I$ to form an attention
mechanism for foreground features and filters away excessive background
noises. The objective can be formulated as:

\vspace{-3mm}

\[
L_{feat}=\frac{1}{2N_{p}}\sum_{l=1}^{L}\sum_{i=1}^{W}\sum_{j=1}^{H}\sum_{c=1}^{C}I_{ij}^{l}\left(f_{adap}^{l}\left(S^{l}\right)_{ijc}-T_{ijc}^{l}\right)^{2}
\]
where $L$, $W$, $H$, $C$ are feature levels, width, height and
number of channels, respectively; $S$ and $T$ are student's and
teacher's feature maps; $I$ is the imitation mask ; $f_{adap}(\cdot)$
is an adaptation function mapping $S$ and $T$ to the same dimension,
which is usually implemented as a 3x3 conv layer; $N_{p}$ is the
total number of positive elements in the mask.

Various methods can differ in the definition of $I$. For example,
FGFI \cite{wang2019distilling} generates an imitation mask according
to the near object anchor locations, while TADF \cite{sun2020distilling}
proposed to generate a soft mask using Gaussian function.

Those mask-based methods are originally proposed for detectors without
FPN. To extend them for the FPN-based modern detectors, the masks
are generated by the same rule for all feature levels, then features
on the corresponding levels are matched for imitation. We argue this
direct adaptation to modern detectors with feature pyramids is suboptimal
due to the following reasons: (1) generally, in the design paradigm
of FPN, each feature level is responsible for detecting objects of
different scales. Thus, indiscriminately applying the same mask on
all levels can introduce noise from unresponsive feature levels; (2)
mask-based methods are not extendible to heterogeneous detector pairs
since the their feature pyramid levels may not be strictly aligned,
e.g., Faster-RCNN-FPN constructs the feature pyramid from $P2$ to
$P6$, while RetinaNet-FPN uses $P3$ to $P7$. These weaknesses promote
us to design a feature imitation mechanism that can automatically
match the features of the student-teacher pair for imitation and eliminate
the excessive noise, while also being extendible to heterogeneous
detector pairs.

\section{Methods}

\subsection{Semantic-Guided Feature Imitation (SGFI)\label{subsec:self-adaptive feature imitation}}

\begin{figure}
\begin{centering}
\vspace{-2mm}
\par\end{centering}
\begin{centering}
\includegraphics[scale=0.81]{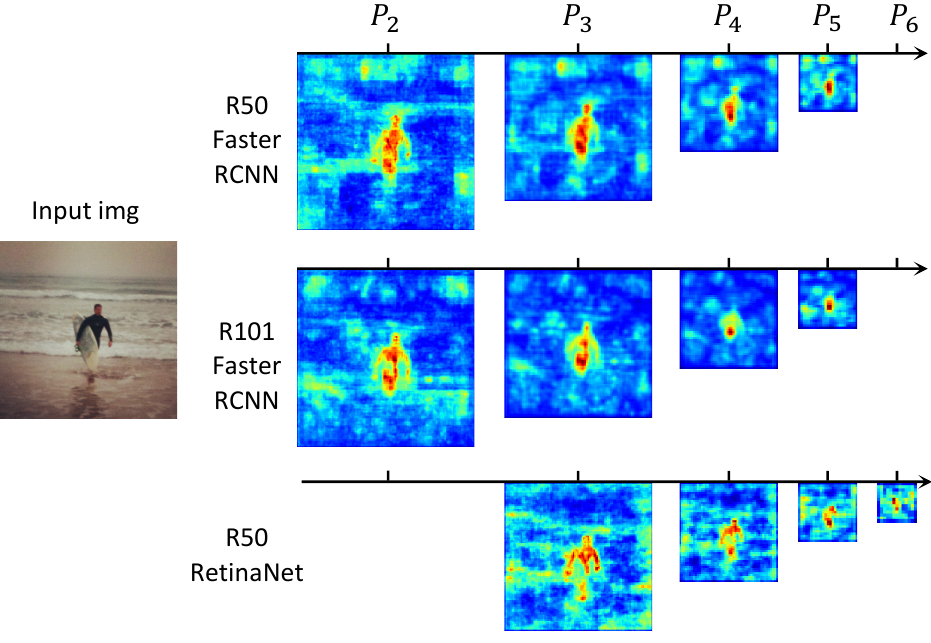}
\par\end{centering}
\caption{\label{fig:featmap}Activation patterns of feature pyramids from different
detectors. We observe that: (1) the same feature level from different
detectors present various patterns; (2) nearby feature maps contain
similar semantics.}
\end{figure}

The aforementioned problem of mask-based methods can be partially
solved via a straightforward approach which uses positive RoI features
for imitation: instead of uniformly imitating foreground regions on
all feature levels, RoI extraction operation selects foreground features
based on object proposals in a fine-grained manner, which automatically
matches the features from corresponding levels according to the their
scales. However, RoI extractor's proposal assignment heuristics is
purely based on the proposal scales, which is agnostic to the semantic
relationship between features of the student-teacher pair.

Intuitively, a promising feature distillation approach needs to consider
the semantics of features when constructing them into pairs for imitation.
We reflect on the feature matching mechanism by looking into the characteristics
of feature pyramids, which are visualized in Figure \ref{fig:featmap}.
We observe that for different detectors, given an object in the image,
the corresponding region on the same feature level present different
patterns, while the difference is more significant between heterogeneous
detectors. This phenomenon implies that the best matching feature
for imitation may not be from the corresponding pyramid level, which
is an issue that the purely scale-based heuristics fails to address.
Thus, we are motivated to conduct feature matching based on their
semantics instead of scales. In addition, the object region on nearby
feature levels are activated in a similar manner, while the similarity
diminishes as the distance between levels becomes greater. This implies
that student features from other levels which carry similar semantics
should also be involved during imitation to fully exploit the potential
of teacher's representation power. 

\begin{figure}
\begin{centering}
\vspace{-2mm}
\par\end{centering}
\begin{centering}
\includegraphics[scale=0.75]{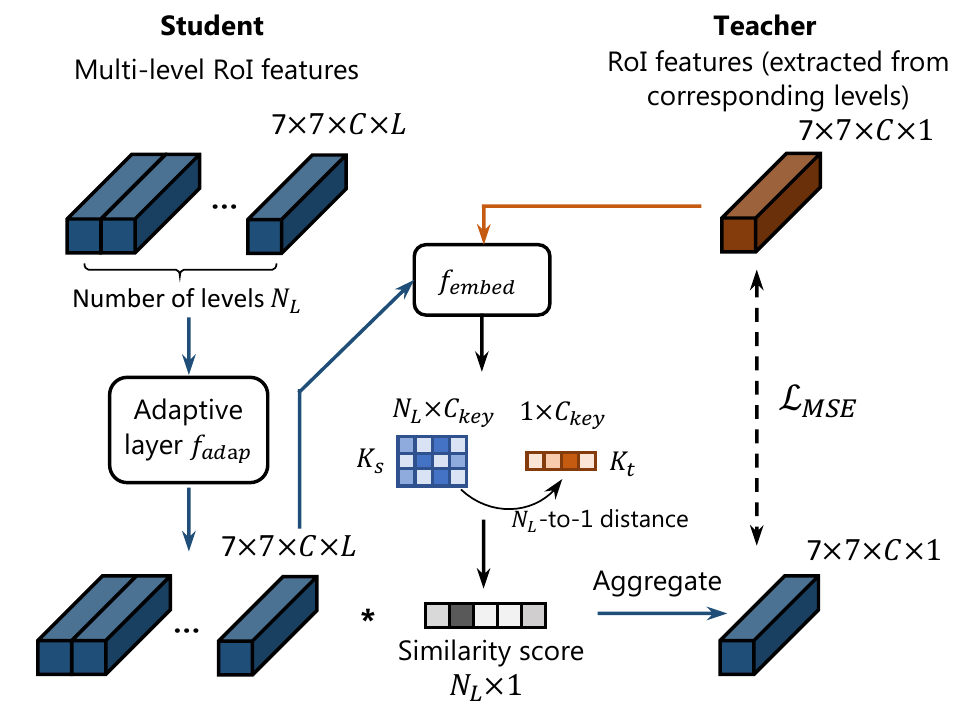}
\par\end{centering}
\caption{\label{fig:cross-level}Framework of our semantic-guided feature imitation.
Our SGFI automatically performs soft matching between feature pairs
across all pyramid levels according to their semantic relations, which
is able to provide the optimal guidance to the student.}

\vspace{-2mm}
\end{figure}

To this end, we propose a semantic-guided feature imitation (\textbf{SGFI})
scheme which performs soft matching between features of the student-teacher
pair as illustrated in Figure \ref{fig:cross-level}. Specifically,
given a proposal indexed by $i$, the teacher's feature $T_{i}\in R^{H\times W\times C}$
is extracted from the heuristically assigned pyramid level (which
is consistent with its training process). In contrast, the student's
features from all levels are extracted and mapped to the same dimension
as $T_{i}$ using $f_{adap}$ and is denoted as $S_{i}\in R^{L\times H\times W\times C}$,
where $L$ is the number of pyramid levels. We first project $T_{i}$
and each $S_{i}^{l}(l=1,...,L)$, onto the same embedding space with
$f_{embed}:R^{H\times W\times C}\rightarrow R^{C_{key}}$, then the
level-wise weights $\alpha_{i}$ are calculated using the dot product
between the embeddings followed by a softmax function, which is used
to aggregate $S_{i}^{l}(l=1,...,L)$ to obtain $S_{agg_{i}}$. The
final loss is the mean square error between $T_{i}$ and $S_{agg_{i}}$.
The calculation of cross-level imitation loss can be formulated as:

\vspace{-5mm}

\[
K_{s_{i}}=f_{embed}\left(f_{adap}\left(S_{i}\right)\right),\;K_{t_{i}}=f_{embed}\left(T_{i}\right)
\]

\vspace{-4mm}

\[
\alpha_{i}=softmax\left(\frac{K_{s_{i}}K_{t_{i}}^{T}}{\tau}\right)
\]

\vspace{-3mm}

\[
S_{agg_{i}}=\sum_{l=1}^{L}\alpha_{i}^{l}\times f_{adap}\left(S_{i}^{l}\right)
\]

\vspace{-3mm}

\begin{equation}
L_{feat}=\frac{1}{N}\sum_{i=1}^{N}\left(MSE\left(S_{agg_{i}},T_{i}\right)\right)\label{eq:featloss}
\end{equation}
where $f_{adapt}$ is implemented as a convolutional layer; $f_{embed}$
is implemented as a lightweight network which consists of 2 convs
with stride=2, each followed by ReLU; both networks for $f_{adapt}$
and $f_{embed}$ are excluded during inference; $N$ and $L$ are
the number of proposals and the number of levels in the feature pyramid,
respectively; $MSE$ is the mean square error; $\tau$ is a learnable
temperature to control the sharpness of softmax logits.

Our proposed SGFI effectively addresses the misalignment between feature
levels of student-teacher pairs, which can be easily extended to heterogeneous
detector pairs.

\subsection{Exploiting Region Relationship with Contrastive KD (CKD)\label{subsec:Contrastive-Knowledge-Distillati}}

Feature imitation methods transfer teacher's knowledge by maximizing
the agreement between features of the same region.  However, the
structural information encoded in the relationship between different
regions is ignored, which may also provide guidance to the student.
To this end, we push the envelop further by integrating the region
relationship into KD for detection. Inspired by the recent work \cite{tian2019contrastive},
we propose to incorporate the idea of contrastive learning into our
KD framework. The objective is to maximize the agreement between the
representations of positive pairs while pushing away those of the
negative pairs in the given metric space, which intrinsically captures
the relationship between contrastive pairs.

Specifically, given a set $B$ consisting of $N$ RoI bounding boxes,
i.e., $B=\{bbox_{i}\}_{i=1,...,N}$, their corresponding representations
$\left\{ r_{s}^{i},r_{t}^{i}\right\} _{i=1,...,N}$ are drawn from
the embeddings before the output layer. We form the contrastive pairs
as follows: representations that correspond to the same box are constructed
as positive pairs while those of different box are constructed as
negatives, namely, $x_{pos}$= $\left\{ r_{s}^{i},r_{t}^{i}\right\} $,
$x_{neg}$=$\left\{ r_{s}^{i},r_{t}^{j}\right\} $$\left(i\neq j\right)$.
Our objective is recognizing the positive pair $x_{pos}$ from the
set $S=\left\{ x_{pos},x_{neg}^{1},x_{neg}^{2},...,x_{neg}^{K}\right\} $
that contains $K$ negative pairs, which is implemented in the form
of InfoNCE loss \cite{oord2018representation}:

\vspace{-3mm}

\begin{equation}
L_{ckd}=\frac{1}{N}\sum_{i=1}^{N}-\log\frac{g\left(r_{s}^{i},r_{t}^{i}\right)}{\sum_{j=0}^{K}g\left(r_{s}^{i},r_{t}^{j}\right)}\label{eq:contrastloss}
\end{equation}
where $K$ is the number of negative samples; $N$ is the number of
proposals in a batch; $g$ is the critic function that estimates the
probability of $\left(r_{s}^{i},r_{t}^{i}\right)$ being the positive
pair, which is defined as:

\vspace{-3mm}

\[
g\left(r_{s},r_{t}\right)=\exp\left(\frac{f_{\theta}\left(r_{s}\right)\cdot f_{\theta}\left(r_{t}\right)}{\parallel f_{\theta}\left(r_{s}\right)\parallel\cdot\parallel f_{\theta}\left(r_{t}\right)\parallel}\cdot\frac{1}{\gamma}\right)
\]
where $f_{\theta}$ is a linear function to project the representation
to a lower dimension, which is implemented as a fully connected layer
whose parameters are shared between the student-teacher pair; $\frac{f_{\theta}\left(r_{s}\right)\cdot f_{\theta}\left(r_{t}\right)}{\parallel f_{\theta}\left(r_{s}\right)\parallel\cdot\parallel f_{\theta}\left(r_{t}\right)\parallel}$
is the cosine similarity; $\gamma$ is a temperature hyper-parameter.

Theoretically, minimizing $L_{ckd}$ is equivalent to maximizing the
lower bound of the mutual information between $f_{\theta}\left(r_{s}\right)$
and $f_{\theta}\left(r_{t}\right)$(Detailed proof can be found in
previous work \cite{oord2018representation}): 

\vspace{-3mm}

\[
MI\left(f_{\theta}\left(r_{s}\right)\cdot f_{\theta}\left(r_{t}\right)\right)\geq\log\left(K\right)-L_{contrastive}
\]

\textbf{Memory Queue.} We implement a memory queue \cite{he2020momentum}
to store the representations for constructing more negative pairs:
a queue across multiple GPUs is maintained, and once the max size
is reached, the oldest batch is dequeued when new batch arrives. Theoretically,
the lower bound becomes tighter as $K$ increases, which implies that
using more negative samples benefits representation learning. However,
we observe that setting $K$ too large leads to performance degradation.
To effect of $K$ is shown in the Appendix.

\textbf{Negative Sample Assignment Strategy. }Another key issue for
contrastive KD in detection task is the mechanism to select negative
samples. Specifically, the dilemma lies in the overlapping between
region proposals: those proposal with large overlaps may contain similar
semantics, thus pushing them away may cause instability during training.
To address this issue, we use IoU to filter out the highly overlapping
proposal boxes and exclude them from negative samples. We conduct
an ablative study in the appendix to decide the optimal IoU threshold.

\subsection{General Detection KD Framework (G-DetKD)}

 In some specific scenarios, only detectors with certain architectures
can be deployed due to hardware constraints, which may cause the student
to have a different architecture from the teacher. Thus, it is promising
if knowledge distillation can be conducted between heterogeneous detector
pairs. However, previous works only consider KD between homogeneous
detectors pairs due to their lack of extensibility. Therefore, we
are motivated to propose a general detection KD framework applicable
for both homogeneous and heterogeneous student-teacher detector pairs.

\subsubsection{Homogeneous Detector Pairs}

\vspace{-1mm}

Homogenous detector pairs are strictly aligned in terms of network
categories and feature representations, which facilitates the design
of KD framework. Other than the previously introduced SGFI and CKD,
we propose two additional techniques to further promote the framework's
effectiveness, namely,\textbf{ class-aware localization KD} and \textbf{head
transfer}. In particular, we study the case when the student and teacher
are both two-stage detectors.

\textbf{Class-aware Localization KD. }A core component distinguishing
detection KD from classification KD lies in how to effectively transfer
teacher's localization ability to student. Intuitively, simply imitating
the four coordinates outputted by the teacher provides limited information
as they are only ``inaccurate targets''. This motivates us to incorporate
teacher's localization knowledge with the class ``uncertainty'',
i.e., utilizing all the class-wise localization predictions generated
by the teacher. For illustration, when the detector captures only
a part of an object, how the box should be shifted may depend on what
class the object belongs to. (This idea requires both student and
teacher to have localization predictions for each possible class,
which is a common setting for two-stage detectors) We calculate the
sum of regression values weighted by classification scores:

\vspace{-3mm}

\begin{equation}
L_{reg}=\frac{1}{N}\sum^{N}\mid\sum_{i=0}^{C}p_{t}^{i}\times\left(reg_{t}^{i}-reg_{s}^{i}\right)\mid\label{eq:reglos}
\end{equation}
where $C$ is the number of classes; $p_{i}$ and $reg_{i}$ are the
classification score and regression outputs of foreground class $i$;
superscripts $s$ and $t$represent student and teacher. This class-aware
regression loss helps student acquire localization knowledge from
teacher; the experiment is shown in Appendix.

\textbf{Head Transfer. }For homogeneous detector pairs, the student
and the teacher have backbones with different capacities, while sharing
the same head structure. This motivates a more straightforward knowledge
transfer strategy for homogeneous detector pairs: directly copying
the weights from the teacher's head to the student. Our experiments
demonstrate it helps accelerate students's convergence and further
improves the performance.

Overall, the loss function of our framework for homogeneous detectors
can be formulated as: $L=L_{gt}+L_{feat}+L_{ckd}+L_{cls}+L_{reg}$,
where $L_{gt}$ is the ground truth loss; $L_{feat}$, $L_{ckd}$
and $L_{reg}$ correspond to \ref{eq:featloss},\ref{eq:contrastloss}
and \ref{eq:reglos}, respectively; $L_{cls}$ is defined in Section
\ref{sec:Preliminary}.

\vspace{-1mm}

\subsubsection{Heterogeneous Detector Pairs}

\begin{figure}
\begin{centering}
\vspace{-2mm}
\par\end{centering}
\begin{centering}
\includegraphics[scale=0.63]{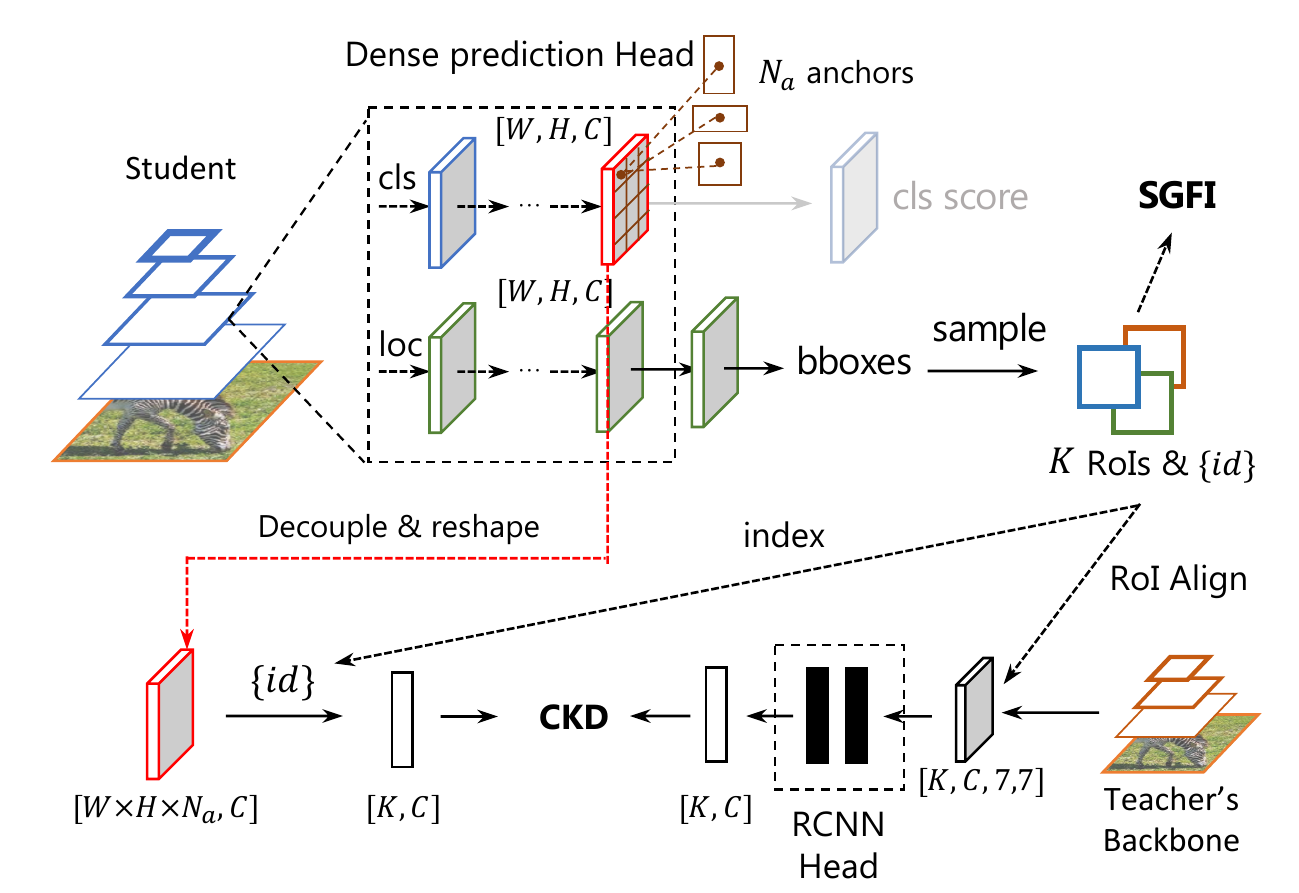}
\par\end{centering}
\vspace{-1mm}
\caption{\label{fig:heterogeneous}Framework of semantic-guided feature distillation
and contrastive distillation for heterogeneous student detectors.
As shown in top of the picture, the proposals are sampled from bounding
boxes predicted by the localization head, which are then used by SGFI
(illustrated in Fig. \ref{fig:cross-level}) and CKD; the bottom part
illustrates the CKD process: student's classification features are
constructed as contrastive pairs with teacher's features. In the case
of one-stage student, each feature is further decoupled for different
box regions centered at the same pixel.}

\vspace{-2mm}
\end{figure}

Modern detectors have diversified into multiple families, such as
one-stage, two-stage and anchor-free detectors. Each family has its
own merits and weaknesses. In particular, two-stage detectors often
have higher performance, while being slower in inference speed. On
the other hand, dense prediction detectors (e.g., one-stage and anchor-free
detectors) are faster than two-stage counterparts while being less
accurate, as they adopt fully convolutional network. In practice,
it is a natural idea to use two-stage detectors as teacher to enhance
detectors belonging to other families.

The difficulty in knowledge distillation between heterogeneous detector
pairs lies in the following aspects: (1) the feature levels are not
aligned, e.g., FasterRCNN constructs the feature pyramid from $P2$
to $P6$, while RetinaNet uses $P3$ to $P7$, which creates obstacle
for feature distillation; (2) different loss function are used during
training, causing their outputs to carry different meanings. E.g.,
two-stage detectors use cross entropy loss, and dense prediction detectors
often adopt focal loss \cite{lin2018focal}, which hinders distillation
on the prediction outputs.

We first try to conduct KD on the prediction outputs of FasterRCNN
teacher and RetinaNet student and find it bringing only limited gains,
and the gains diminish when applied with other KD methods (the results
are shown in Table \ref{tab:retina-pred}). Next, we elaborate on
how to extend our feature level distillation method to heterogeneous
detectors.

The overview of our G-DetKD for heterogeneous detector pairs is shown
in Figure \ref{fig:heterogeneous}. We use the student's bounding
box predictions to generate a set of RoIs, which are then applied
to cross-level feature imitation and contrastive KD in the same way
as described in Sections \ref{subsec:self-adaptive feature imitation}
and \ref{subsec:Contrastive-Knowledge-Distillati}.

\textbf{RoI Extraction. }Analogous to the RPN head in two-stage detectors,
we first match all the student's predicted boxes with the ground truths
and label those with IoUs greater than a threshold (set to be 0.5)
as positive samples. Then the boxes are sampled with 1:3 ratio of
positive to negative samples.

\textbf{Semantic-guided Feature Imitation. }As introduced in \ref{subsec:self-adaptive feature imitation},
our method eliminates the requirement for strict alignment between
feature levels of the student-teacher pairs, thus is directly applicable
to heterogeneous detector pairs.

\textbf{Decoupling Representations for Contrastive KD}. The typical
bounding box head structures of dense prediction detector consists
of separate multi-convolution classification and localization branches.
As illustrated in Figure \ref{fig:heterogeneous}, given a set of
RoIs, the contrastive pairs are constructed using the features of
the teacher's last fully connected layer and the corresponding features
from the last layer of student's classification branch. For one-stage
detectors, each representation encodes information of multiple anchors
centered at the same location, which is decoupled for each anchor.

Overall, the loss function of our framework for heterogeneous detectors
can be formulated as: $L=L_{gt}+L_{feat}+L_{ckd}$, where $L_{gt}$
is the ground truth loss; $L_{feat}$ and $L_{ckd}$ correspond to
\ref{eq:featloss},\ref{eq:contrastloss} and \ref{eq:reglos}, respectively.

\begin{table}
\vspace{-2mm}

\begin{centering}
{\footnotesize{}}%
\begin{tabular}{c|c}
\hline 
{\footnotesize{}Part} & \multicolumn{1}{c}{{\footnotesize{}AP}}\tabularnewline
\hline 
{\footnotesize{}Baseline R18} & {\footnotesize{}32.3}\tabularnewline
\hline 
{\footnotesize{}Cls pred} & {\footnotesize{}32.6$^{+0.3}$}\tabularnewline
{\footnotesize{}Reg pred} & {\footnotesize{}32.7$^{+0.4}$}\tabularnewline
\textbf{\footnotesize{}SGFI} & \textbf{\footnotesize{}36.1$^{+3.8}$}\tabularnewline
{\footnotesize{}SGFI+pred(cls+reg)} & {\footnotesize{}36.1$^{+3.8}$}\tabularnewline
\hline 
\end{tabular}{\footnotesize\par}
\par\end{centering}
\vspace{1mm}

\caption{\label{tab:retina-pred}Knowledge distillation between R18-RetinaNet
student and R50-fasterRCNN teacher. The gain of prediction KD is negligible
and diminishes when applied with our SGFI.}
\vspace{-2mm}
\end{table}

\section{Experiments}

\textbf{Datasets.} We evaluate our knowledge distillation framework
on various modern object detectors and popular benchmarks. Our main
experiments are conducted on COCO dataset \cite{Lin2014}. When compared
with other popular algorithms, \texttt{test-dev} split is used and
the performances are obtained by uploading the results to the COCO
test server. \textbf{Berkeley Deep Drive} (BDD) \cite{Yu2018} and
\textbf{PASCAL VOC} (VOC) \cite{Everingham10} are then used to validate
the generalization capability of our method. The default evaluation
metrics for each dataset is adopted.

\textbf{Implementation Detail. }We use cosine annealing learning rate
schedule; the initial learning rate is set to 0.03. Training is conducted
on 8 GPUs using synchronized SGD, batch size is 2 for each GPU. The
shorter side of the input image is scaled to 800 pixels, the longer
side is limited up to 1333 pixels. '1x' (namely 12 epochs) and '2x+ms'
(namely 24 epochs with multi-scale training) training schedules are
used\textbf{.} More details can be found in Appendix.\textbf{ }

\subsection{Main Results}

\textbf{Decoupling the effectiveness of each distillation component}.
We first investigate the effectiveness of each distillation component
for three types of student detectors. The experiment results are shown
in Table \ref{tab:KD component}, where use FasterRCNN-R50-FPN as
teacher and all students have R18-FPN backbones. The result for FasterRCNN
student demonstrates the effectiveness of our class-aware localization
KD, which outperforms the naive localization KD approach by a large
margin. In addition, contrastive KD is the most effective approach
when used alone, which leads to $3.1\%$ gain in $AP$. On the other
hand, for heterogeneous students, we can see our SGFI and Contrastive
KD provide even higher performance gain compared to their results
for homogeneous pairs. The combination achieving the highest performance
gain is selected for other main experiments.

\begin{table}
\vspace{-1mm}

\begin{centering}
\tabcolsep 0.01in%
\begin{tabular}{l|c|c|c}
\hline 
\multirow{2}{*}{{\footnotesize{}KD Method}} & \multicolumn{3}{c}{{\footnotesize{}AP}}\tabularnewline
\cline{2-4} \cline{3-4} \cline{4-4} 
 & {\footnotesize{}FasterRCNN\cite{Ren2015}} & {\footnotesize{}Retina\cite{lin2018focal}} & {\footnotesize{}FCOS\cite{tian2019fcos}}\tabularnewline
\hline 
{\footnotesize{}Teacher R50-FPN (2x+ms)} & {\footnotesize{}39.9} & {\footnotesize{}-} & {\footnotesize{}-}\tabularnewline
{\footnotesize{}Student R18-FPN} & {\footnotesize{}34.0} & {\footnotesize{}32.6} & {\footnotesize{}30.3}\tabularnewline
\hline 
{\footnotesize{}SGFI} & {\footnotesize{}36.7$^{+2.7}$} & {\footnotesize{}36.0$^{+3.4}$} & {\footnotesize{}35.2$^{+4.9}$}\tabularnewline
{\footnotesize{}CKD} & {\footnotesize{}$^{\dagger}$37.1$^{+3.1}$} & {\footnotesize{}35.8$^{+3.2}$} & {\footnotesize{}34.9$^{+4.6}$}\tabularnewline
{\footnotesize{}Pred (Cls+Reg)} & {\footnotesize{}36.4$^{+2.4}$} & {\footnotesize{}33.1$^{+0.5}$} & {\footnotesize{}-}\tabularnewline
{\footnotesize{}Pred (Cls+CAReg)} & {\footnotesize{}37.0$^{+3.0}$} & {\footnotesize{}-} & -\tabularnewline
{\footnotesize{}HT} & {\footnotesize{}35.3$^{+1.3}$} & {\footnotesize{}-} & {\footnotesize{}-}\tabularnewline
\textbf{\footnotesize{}SGFI+CKD} & {\footnotesize{}37.6$^{+3.6}$} & {\footnotesize{}36.3$^{+3.7}$} & {\footnotesize{}35.6$^{+5.3}$}\tabularnewline
\textbf{\footnotesize{}SGFI+CKD+Pred+HT} & {\footnotesize{}37.9$^{+3.9}$} & - & -\tabularnewline
\hline 
\end{tabular}
\par\end{centering}
\vspace{1mm}

\caption{\label{tab:KD component}Effectiveness of each KD component for different
type of students. SGFI and CKD are our semantic-guided feature imitation
and contrastive KD, respectively; Cls and CAReg are classification
KD and our class-aware regression KD, respectively; HT is our head
transfer technique.}

\vspace{-1mm}
\end{table}

\textbf{Comparison with SOTA Detection KD Methods. }We compare our
results for homogeneous pairs with previous works in Table \ref{tab:comparison_with_previous-1},
since those methods only consider distillation between homogeneous
teacher-student pairs. The results show that our G-DetKD and CKD consistently
outperforms others by a large margin.

\begin{table}
\vspace{-1mm}

\begin{centering}
{\footnotesize{}}%
\begin{tabular}{l|cccc}
\hline 
{\footnotesize{}Method} & {\footnotesize{}AP} & {\footnotesize{}${\rm AP_{S}}$} & {\footnotesize{}${\rm AP_{M}}$} & {\footnotesize{}${\rm AP_{L}}$}\tabularnewline
\hline 
{\footnotesize{}Teacher R152-FPN} & {\footnotesize{}41.5} & {\footnotesize{}24.1} & {\footnotesize{}45.8} & {\footnotesize{}54.0}\tabularnewline
{\footnotesize{}Student R50-FPN} & {\footnotesize{}37.4} & {\footnotesize{}21.8} & {\footnotesize{}41.0} & {\footnotesize{}47.8}\tabularnewline
{\footnotesize{}FGFI\cite{wang2019distilling}} & {\footnotesize{}39.8$^{+2.4}$} & {\footnotesize{}22.9} & {\footnotesize{}43.6} & {\footnotesize{}52.8}\tabularnewline
{\footnotesize{}TADF\cite{sun2020distilling}} & {\footnotesize{}40.0$^{+2.6}$} & {\footnotesize{}23.0} & {\footnotesize{}43.6} & {\footnotesize{}53.0}\tabularnewline
\textbf{\footnotesize{}CKD (Ours)} & \textbf{\footnotesize{}40.3$^{+2.9}$} & \textbf{\footnotesize{}23.2} & \textbf{\footnotesize{}44.1} & \textbf{\footnotesize{}52.3}\tabularnewline
\textbf{\footnotesize{}G-DetKD (Ours)} & \textbf{\footnotesize{}41.0$^{+3.6}$} & \textbf{\footnotesize{}23.7} & \textbf{\footnotesize{}45.0} & \textbf{\footnotesize{}53.7}\tabularnewline
\hline 
{\footnotesize{}Teacher R50-FPN} & {\footnotesize{}37.4} & {\footnotesize{}21.8} & {\footnotesize{}41.0} & {\footnotesize{}47.8}\tabularnewline
{\footnotesize{}Student R50(1/4)-FPN} & {\footnotesize{}29.4} & {\footnotesize{}16.3} & {\footnotesize{}31.7} & {\footnotesize{}39.0}\tabularnewline
{\footnotesize{}FGFI\cite{wang2019distilling}} & {\footnotesize{}31.7$^{+2.3}$} & {\footnotesize{}17.1} & {\footnotesize{}34.2} & {\footnotesize{}43.0}\tabularnewline
\textbf{\footnotesize{}CKD (Ours)} & \textbf{\footnotesize{}32.4$^{+3.0}$} & \textbf{\footnotesize{}17.2} & \textbf{\footnotesize{}34.8} & \textbf{\footnotesize{}43.0}\tabularnewline
\textbf{\footnotesize{}G-DetKD (Ours)} & \textbf{\footnotesize{}33.7$^{+4.3}$} & \textbf{\footnotesize{}18.1} & \textbf{\footnotesize{}36.6} & \textbf{\footnotesize{}44.5}\tabularnewline
\hline 
\end{tabular}{\footnotesize\par}
\par\end{centering}
\vspace{1mm}

\caption{\label{tab:comparison_with_previous-1}Comparison between our KD methods
with other approaches. The results show that our G-DetKD and CKD consistently
outperforms others by a large margin.}

\vspace{-1mm}
\end{table}

\textbf{Exploit the Full Potential of Whole Framework.} In order
to further explore the full potential of our G-DetKD, we use a more
powerful X101-FasterRCNN-InstaBoost model as the teacher and train
the student for using 2x+ms schedule. We conduct experiments on students
belonging to four families. The results shown in Table \ref{tab:Main-results-1}
demonstrate the significant performance gains brought by our method:
$+3.8$ for R101-RetinaNet, $+3.1$ for R101-FCOS, $+3.4$ for R101-FasterRCNN
and $+4.1$ for X101-Cascade (with X101-HTC teacher). In addition,
we compare the performance of our KD-enhanced detectors with the current
SOTA detector designs on COCO \texttt{test-dev} split in Table \ref{tab:Main-exp}.
The results show that our KD method can upgrade regular detectors
to easily dominate the SOTA methods without any modification to the
detector. Since KD and detector design are parallel approaches, there
is potential for further improvement if we apply our KD framework
to the best detector designs.

\begin{table}
\begin{centering}
\tabcolsep 0.01in{\footnotesize{}}%
\begin{tabular}{l|c|c|c}
\hline 
\multicolumn{2}{c|}{{\footnotesize{}Student}} & {\footnotesize{}Teacher} & {\footnotesize{}AP}\tabularnewline
\hline 
{\footnotesize{}R18 (34.2)} & \multirow{3}{*}{{\footnotesize{}RetinaNet\cite{lin2018focal}}} & \multirow{3}{*}{} & {\footnotesize{}38.5$^{+4.3}$}\tabularnewline
{\footnotesize{}R50 (38.5)} &  &  & {\footnotesize{}43.3$^{+4.8}$}\tabularnewline
{\footnotesize{}R101 (40.8)} &  &  & {\footnotesize{}44.6$^{+3.8}$}\tabularnewline
\cline{1-2} \cline{2-2} \cline{4-4} 
{\footnotesize{}R18 (34.4)} & \multirow{3}{*}{{\footnotesize{}FCOS\cite{tian2019fcos}}} & {\footnotesize{}X101} & {\footnotesize{}38.6$^{+4.2}$}\tabularnewline
{\footnotesize{}R50 (38.6)} &  & {\footnotesize{}FasterRCNN} & {\footnotesize{}43.1$^{+4.5}$}\tabularnewline
{\footnotesize{}R101 (41.5)} &  & {\footnotesize{}InstaBoost\cite{fang2019instaboost}(44.5)} & {\footnotesize{}44.6$^{+3.1}$}\tabularnewline
\cline{1-2} \cline{2-2} \cline{4-4} 
{\footnotesize{}R18 (36.0)} & \multirow{3}{*}{{\footnotesize{}FasterRCNN\cite{Ren2015}}} & \multirow{3}{*}{} & {\footnotesize{}39.9$^{+3.9}$}\tabularnewline
{\footnotesize{}R50 (39.9)} &  &  & {\footnotesize{}44.0$^{+4.1}$}\tabularnewline
{\footnotesize{}R101 (41.8)} &  &  & {\footnotesize{}45.2$^{+3.4}$}\tabularnewline
\hline 
\multirow{2}{*}{{\footnotesize{}X101-Cascade(46.0)}} & \multirow{2}{*}{{\footnotesize{}CascadeRCNN\cite{cai2017cascade}}} & \multirow{2}{*}{{\footnotesize{}X101-HTC\cite{chen2019hybrid}(50.4)}} & \multirow{2}{*}{\textbf{\footnotesize{}50.1}{\footnotesize{}$^{+4.1}$}}\tabularnewline
 &  &  & \tabularnewline
\hline 
\end{tabular}{\footnotesize\par}
\par\end{centering}
\vspace{1mm}

\caption{\label{tab:Main-results-1}Our detection KD framework brings significant
performance boosts for both homogeneous and heterogeneous detector
pairs. All students use two-stage detectors as teachers. \textquotedblleft Insta\textquotedblright{}
means Instaboost \cite{fang2019instaboost}; \textquotedblleft HTC\textquotedblright{}
stands for Hybrid Task Cascade \cite{chen2019hybrid}.}
\end{table}

\begin{table}
\vspace{-2mm}

\begin{centering}
{\small{}\tabcolsep 0.02in}%
\begin{tabular}{l|c|ccc}
\hline 
{\footnotesize{}Method} & {\footnotesize{}backbone} & {\footnotesize{}${\rm AP}$} & {\footnotesize{}${\rm AP_{@.5}}$} & {\footnotesize{}${\rm AP_{@.7}}$}\tabularnewline
\hline 
{\footnotesize{}RepPoints \cite{yang2019reppoints}} & {\footnotesize{}R101-DCN} & {\footnotesize{}45.0} & {\footnotesize{}66.1} & {\footnotesize{}49.0}\tabularnewline
{\footnotesize{}SAPD \cite{zhu2019soft}} & {\footnotesize{}R101} & {\footnotesize{}43.5} & {\footnotesize{}63.6} & {\footnotesize{}46.5}\tabularnewline
{\footnotesize{}ATSS \cite{zhang2020bridging}} & {\footnotesize{}R101} & {\footnotesize{}43.6} & {\footnotesize{}62.1} & {\footnotesize{}47.4}\tabularnewline
{\footnotesize{}PAA \cite{kim2020probabilistic}} & {\footnotesize{}R101} & {\footnotesize{}44.8} & {\footnotesize{}63.3} & {\footnotesize{}48.7}\tabularnewline
{\footnotesize{}BorderDet \cite{qiu2020borderdet}} & {\footnotesize{}R101} & {\footnotesize{}45.4} & {\footnotesize{}64.1} & {\footnotesize{}48.8}\tabularnewline
{\footnotesize{}BorderDet \cite{qiu2020borderdet}} & {\footnotesize{}X101-64x4d-DCN} & {\footnotesize{}47.2} & {\footnotesize{}66.1} & {\footnotesize{}51.0}\tabularnewline
\hline 
{\footnotesize{}RetinaNet (Ours)} & {\footnotesize{}R101} & \textbf{\footnotesize{}44.8} & \textbf{\footnotesize{}64.2} & \textbf{\footnotesize{}48.3}\tabularnewline
{\footnotesize{}FCOS (Ours)} & {\footnotesize{}R101} & \textbf{\footnotesize{}45.0} & \textbf{\footnotesize{}64.1} & \textbf{\footnotesize{}48.5}\tabularnewline
{\footnotesize{}FasterRCNN (Ours)} & {\footnotesize{}R101} & \textbf{\footnotesize{}45.6} & \textbf{\footnotesize{}65.9} & \textbf{\footnotesize{}49.9}\tabularnewline
{\footnotesize{}Cascade (Ours)} & {\footnotesize{}X101-32x4d-DCN} & \textbf{\footnotesize{}50.5} & \textbf{\footnotesize{}69.3} & \textbf{\footnotesize{}55.1}\tabularnewline
\hline 
\end{tabular}{\small\par}
\par\end{centering}
\vspace{1mm}

\caption{\label{tab:Main-exp}Comparison of our G-DetKD with SOTA detector
design approaches. Detectors in the four last rows are enhanced by
our G-DetKD framework.}
\vspace{-1mm}
\end{table}

\subsection{Ablation Study\label{subsec:Ablative-Study}}

\textbf{Comparison of Feature Imitation Strategies.} We evaluate different
feature imitation strategies, including various mask-based methods,
RoI feature imitation (RoIFI) and our SGFI, for both homogeneous and
heterogeneous student-teacher pairs in Table \ref{tab:feature imitation}.
For homogeneous detectors, mask-based approaches show similar performances,
which are outperformed by RoI-based imitation methods by a large margin.
Our proposed SGFI beats the best mask-based approach by $1.1\%$ AP.
For heterogeneous detectors, the superiority of SGFI becomes more
evident, which indicates that SGFI can well resolve the misalignment
issue between feature levels of the student-teacher pairs.

\begin{table}
\vspace{-2mm}

\begin{centering}
{\scriptsize{}}%
\begin{tabular}{c|c|c|c}
\hline 
\multirow{2}{*}{{\footnotesize{}Method}} & \multicolumn{3}{c}{{\footnotesize{}AP}}\tabularnewline
\cline{2-4} \cline{3-4} \cline{4-4} 
 & {\footnotesize{}FasterRCNN} & {\footnotesize{}Retina} & {\footnotesize{}FCOS}\tabularnewline
\hline 
{\footnotesize{}Teacher R50-FPN (2x+ms)} & {\footnotesize{}39.9} & {\footnotesize{}-} & {\footnotesize{}-}\tabularnewline
{\footnotesize{}Student R18-FPN} & {\footnotesize{}34.0} & {\footnotesize{}32.6} & {\footnotesize{}30.3}\tabularnewline
\hline 
{\footnotesize{}Whole Feature \cite{chen2017learning}} & {\footnotesize{}35.2$^{+1.2}$} & {\footnotesize{}-} & {\footnotesize{}-}\tabularnewline
{\footnotesize{}Anchor Mask \cite{wang2019distilling}} & {\footnotesize{}35.6$^{+1.6}$} & {\footnotesize{}-} & {\footnotesize{}-}\tabularnewline
{\footnotesize{}Gaussian Mask \cite{sun2020distilling}} & {\footnotesize{}35.4$^{+1.4}$} & {\footnotesize{}-} & {\footnotesize{}-}\tabularnewline
{\footnotesize{}GT Mask} & {\footnotesize{}35.5$^{+1.5}$} & {\footnotesize{}34.3$^{+1.7}$} & {\footnotesize{}31.8$^{+1.5}$}\tabularnewline
\hline 
{\footnotesize{}RoIFI} & {\footnotesize{}36.4 $^{+2.4}$} & {\footnotesize{}35.5$^{+2.9}$} & {\footnotesize{}34.7$^{+4.4}$}\tabularnewline
\textbf{\footnotesize{}SGFI} & \textbf{\footnotesize{}36.7$^{+2.7}$} & \textbf{\footnotesize{}36.0$^{+3.4}$} & {\footnotesize{}35.2$^{+4.9}$}\tabularnewline
\hline 
\end{tabular}{\scriptsize\par}
\par\end{centering}
\vspace{1mm}

\caption{\label{tab:feature imitation}Comparison between different feature
distillation techniques. As can be seen, {\small{}our SGFI} outperforms
the other methods by a large margin. RoIFI stands for RoI feature
imitation method.}
\vspace{-2mm}
\end{table}

\begin{figure}
\begin{centering}
\vspace{-2mm}
\par\end{centering}
\begin{centering}
\includegraphics[scale=0.29]{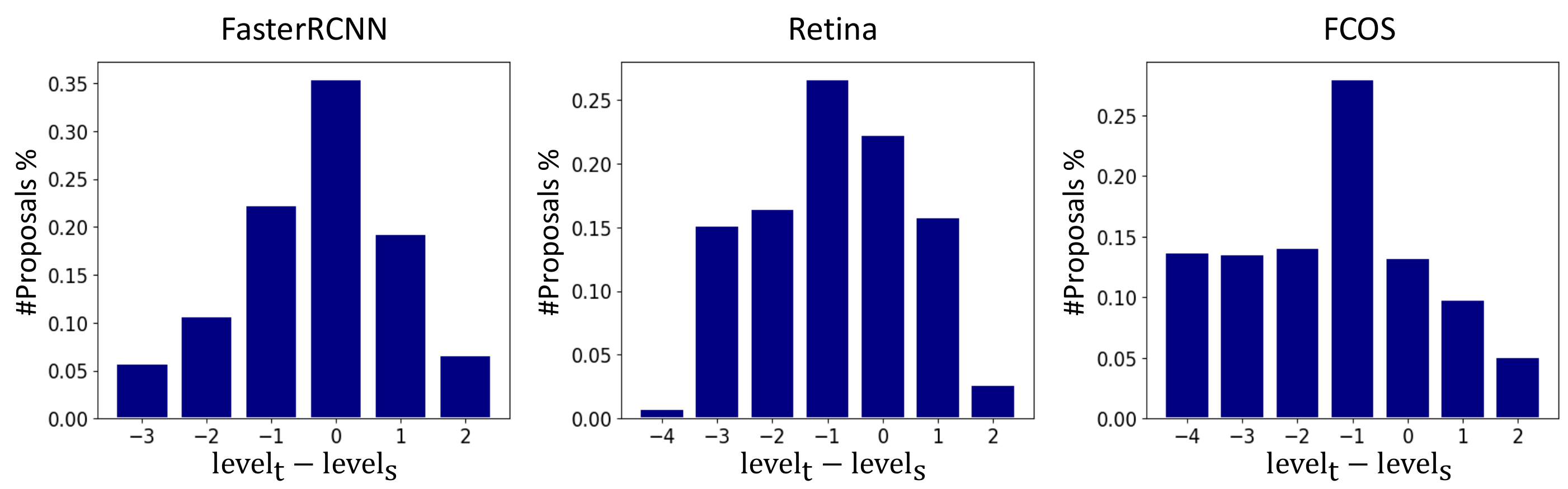}
\par\end{centering}
\caption{\label{fig:feat_level_diff}Distributions of the pyramid level difference
between the student-teacher pair's best matching features.}

\vspace{-1mm}
\end{figure}

\textbf{Visualizing Pyramid Level Matching in SGFI.} To investigate
the feature's matching pattern in SGFI, we visualize the distributions
of the pyramid level difference between the student-teacher pair's
best matching features in Figure \ref{fig:feat_level_diff}. Specifically,
500 images (consisting of around 50000 positive proposals) are fed
into the trained student-teacher pair. For each proposal, we collect
its corresponding teacher's pyramid level and student's best matching
level to calculate the difference. The distributions show that, while
most of the best matching features are from the same level, feature
matchings from different levels exist. In addition, the possibility
of matching diminishes as features' corresponding levels are farther
apart. Note that the starting index of pyramids in RetinaNet and FCOS
is larger than that of the teacher by 1, as explained in \ref{sec:Preliminary},
thus causing the distribution to be centered at -1.

\begin{figure}
\begin{centering}
\vspace{-1mm}
\par\end{centering}
\begin{centering}
\includegraphics[scale=0.25]{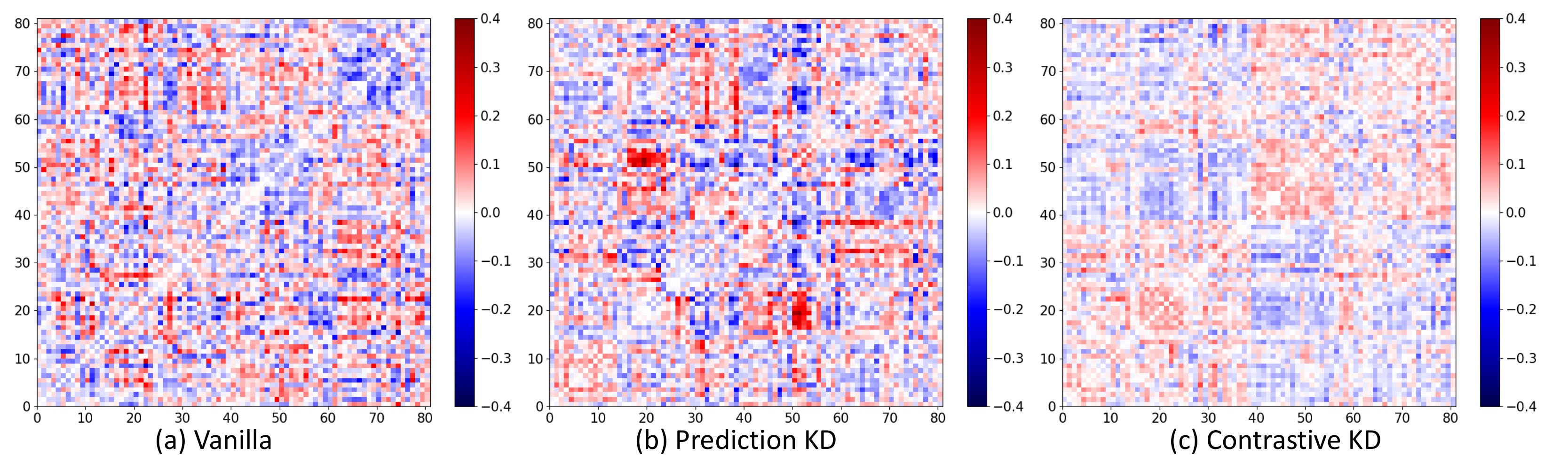}
\par\end{centering}
\caption{\label{fig:corr-1-1-1}Visualization of the difference between the
student-teacher pair's correlation matrices at the classification
logits. The intensity of the color represents the magnitude.}

\vspace{-1mm}
\end{figure}

\textbf{Visualizing the Influence of} \textbf{Contrastive KD at Class
Outputs.} To verify the effectiveness of our contrastive KD, we visualize
the difference between the correlation matrices of the student-teacher
pair's classification logits. As can be seen in fig \ref{fig:corr-1-1-1},
in comparison with baseline training and prediction KD strategy, the
resulting student model of our contrastive KD approach has higher
correlation in the classification outputs with the teacher, which
indicates that our contrastive KD effectively captures the inter-class
correlations and helps the student learn the complex structural information
(i.e., the interdependencies between output dimensions) from the teacher.

\textbf{Generalization Ability of our KD framework.} The generalization
ability is important for training strategies, for which we explore
from two aspects: model and dataset. Homogeneous detector pairs are
used throughout the experiments.

\textbf{Generalization for Students and Teachers.} A robust KD framework
should generalize for models with different capacities. We explore
our G-DetKD's generalization ability in Table \ref{tab:capacity-generalization}.
The results show that our method consistently results in performance
gains for student-teacher pairs with different capacities. Interestingly,
we find that student's performance is boosted even when the teacher
is weaker. We assume this phenomenon is due to the regularization
effect of our KD method, which is also observed in \cite{yuan2020revisiting}.

\begin{table}
\begin{centering}
{\footnotesize{}}%
\begin{tabular}{c|c|c}
\hline 
{\footnotesize{}Student} & {\footnotesize{}Teacher} & {\footnotesize{}AP}\tabularnewline
\hline 
{\footnotesize{}R18 (34.0)} & {\footnotesize{}R50 (39.9)} & {\footnotesize{}37.9$^{+3.9}$}\tabularnewline
{\footnotesize{}R50 (37.4)} & {\footnotesize{}R50 (39.9)} & {\footnotesize{}39.8$^{+2.4}$}\tabularnewline
{\footnotesize{}R101 (39.4)} & {\footnotesize{}R50 (39.9)} & {\footnotesize{}40.4$^{+1.0}$}\tabularnewline
\hline 
{\footnotesize{}R50 (37.4)} & {\footnotesize{}R18 (36.0)} & {\footnotesize{}37.7$^{+0.3}$}\tabularnewline
{\footnotesize{}R50 (37.4)} & {\footnotesize{}R50 (39.9)} & {\footnotesize{}39.8$^{+2.4}$}\tabularnewline
{\footnotesize{}R50 (37.4)} & {\footnotesize{}R101 (41.8)} & {\footnotesize{}41.1$^{+3.7}$}\tabularnewline
\hline 
\end{tabular}{\footnotesize\par}
\par\end{centering}
\vspace{1mm}

\caption{\label{tab:capacity-generalization} Our KD framework consistently
boosts the performance given students and teachers with different
capacities. The values in the parentheses indicate baseline APs.}
\vspace{-2mm}
\end{table}

\textbf{Generalization on other Datasets.}\label{Generalization-on-other-1}
We show our method generalizes well for different datasets by experimenting
on two additional datasets, namely Pascal VOC \cite{Everingham10}
and BDD \cite{Yu2018}. We do not deliberately pick powerful teachers
(R101-FPN are used as teachers and is trained under the same schedule
as the student). Our R50-FPN achieves 3.8\% gain in ${\rm AP_{@.9}}$
on VOC and 1.4\% {\footnotesize{}${\rm AP}$ }gain on BDD.

\begin{table}[h]
\vspace{-2mm}

\begin{centering}
{\footnotesize{}}%
\begin{tabular}{c|ccc}
\hline 
{\footnotesize{}Model} & {\footnotesize{}$\mathrm{{\textstyle AP_{@.5}}}$} & {\footnotesize{}${\rm AP_{@.7}}$} & {\footnotesize{}${\rm AP_{@.9}}$}\tabularnewline
\hline 
{\footnotesize{}R50-FPN} & {\footnotesize{}81.8$\uparrow$83.0} & {\footnotesize{}67.5$\uparrow$69.7} & {\footnotesize{}16.8$\uparrow$20.6}\tabularnewline
{\footnotesize{}R101-FPN} & {\footnotesize{}83.3$\uparrow$83.7} & {\footnotesize{}69.7$\uparrow$70.3} & {\footnotesize{}19.8$\uparrow$22.4}\tabularnewline
\hline 
\end{tabular}{\footnotesize\par}
\par\end{centering}
\begin{centering}
{\footnotesize{}(a) VOC}{\footnotesize\par}
\par\end{centering}
\begin{centering}
{\footnotesize{}}%
\begin{tabular}{c|cccc}
\hline 
{\footnotesize{}Model} & {\footnotesize{}${\rm AP}$} & {\footnotesize{}${\rm AP_{S}}$} & {\footnotesize{}${\rm AP_{M}}$} & {\footnotesize{}${\rm AP_{L}}$}\tabularnewline
\hline 
{\footnotesize{}R50-FPN} & {\footnotesize{}36.5$\uparrow$37.9} & {\footnotesize{}14.5$\uparrow$14.6} & {\footnotesize{}39.0$\uparrow$45.0} & {\footnotesize{}56.9$\uparrow$58.5}\tabularnewline
{\footnotesize{}R101-FPN} & {\footnotesize{}37.5$\uparrow$38.0} & {\footnotesize{}14.5$\uparrow$14.7} & {\footnotesize{}40.0$\uparrow$40.6} & {\footnotesize{}57.7/$\uparrow$58.4}\tabularnewline
\hline 
\end{tabular}{\footnotesize\par}
\par\end{centering}
\begin{centering}
{\footnotesize{}(b) BDD}{\footnotesize\par}
\par\end{centering}
\caption{\label{tab:VOC and BDD-1}Generalization on VOC and BDD datasets.
The value at the left of {\footnotesize{}$\uparrow$ }is the baseline
AP while the value at the right is AP achieved with KD.}
\vspace{-2mm}
\end{table}

\section{Conclusion}

In this paper, we propose a semantic-guided feature imitation method
and contrastive distillation for detectors, which helps the student
better exploit the learnt knowledge from teacher's feature pyramid.
Furthermore, a general KD framework for detectors is proposed, which
is applicable for both homogeneous and heterogeneous detector pairs.

{\small{}\bibliographystyle{ieee_fullname}
\bibliography{reference}
}{\small\par}

\begin{appendix}

\part*{Supplementary Materials}

\section{Experiment Setups}

\textbf{Datasets and evaluation metrics.} We evaluate our knowledge
distillation framework on various modern object detectors and popular
benchmarks. Our main experiments are conducted on COCO dataset \cite{Lin2014}:
we use \texttt{train2017} split (115K images) to perform training
and validate the result on \texttt{minival} split (5K images). When
compared with other popular algorithms, \texttt{test-dev} split is
used and the performances are obtained by uploading the results to
the COCO test server. \textbf{Berkeley Deep Drive} (BDD) \cite{Yu2018}
and \textbf{PASCAL VOC} (VOC) \cite{Everingham10} are then used to
validate the generalization capability of our method: BDD is an autonomous
driving dataset containing 10 object classes, 70K images for training
and 10K for evaluation; for VOC, \texttt{trainval07} split is used
for training and \texttt{test2007} split is used for testing. For
experiments on COCO and BDD, mAP for IoU thresholds from 0.5 to 0.95
is used as the performance metric; while AP at IOU = 0.5 is used for
VOC.

\textbf{Additional Implementation details.} Other than the settings
mentioned in the main paper's Main Results section, we provide the
additional details as follows: weight decay is set to 0.0001, momentum
is set to 0.9. For contrastive KD, $K=80*1024$ (1024 proposals per
GPU for 8 GPUs) proposals are used to form the memory queue; When
Transfer Head is applied, the weights of transferred RPN and RCNN
are frozen throughout the training process. The checkpoints of all
teacher models in following sections can be easily obtained from the
MMDetection \cite{mmdetection2018} official website\footnote{https://github.com/open-mmlab/mmdetection}.

\section{Contrastive KD Implementation}

In this section, we carefully analyze the influence of the important
factors in our contrastive KD. As introduced in Section 4.2 of the
main paper, the size of memory queue and the IoU threshold for assigning
negative samples both play important roles in the performance of our
contrastive KD (CKD). Thus, we conduct ablative experiments to find
the optimal values for those hyper-parameters. For those experiments,
R50-FasterRCNN is used as the teacher, while R18-FasterRCNN is used
as the student. The training schedule is 1x.

\textbf{Memory Queue.} We implement a memory queue borrowing the idea
from \cite{he2020momentum} to increase the number of negative samples.
As given by the \textbf{Proofs}, a large queue size $K$ is theoretically
beneficial for the training objective. However, we observe that a
large $K$ does not necessarily lead to a better result. The optimal
value for $K$ is around 80x1024. As a single GPU has 1024 proposal
features per batch, 8x1024 proposal features can be directly obtained
by gathering from all 8 GPUs, the additional negative samples are
formed using representations from previous batches. In addition, we
observe that the training becomes unstable and the loss often explodes
when $K$ is too large. The experimental results are shown in Table
\ref{tab:banksize-1}.

\textbf{IoU Threshold. }In object detection, multiple proposals may
be overlapping with each other, forming those proposal representations
with similar semantics into negative pairs and forcing them to be
apart is suboptimal. To address this issue, we use IoU to filter out
the highly overlapping proposal boxes and exclude them from negative
samples. We conduct experiments to decide the optimal IoU threshold,
the results are shown in Table \ref{tab:IoUthreshold}. We observe
that the best threshold value is around 0.5.

\begin{figure}
\begin{centering}
\includegraphics[scale=0.35]{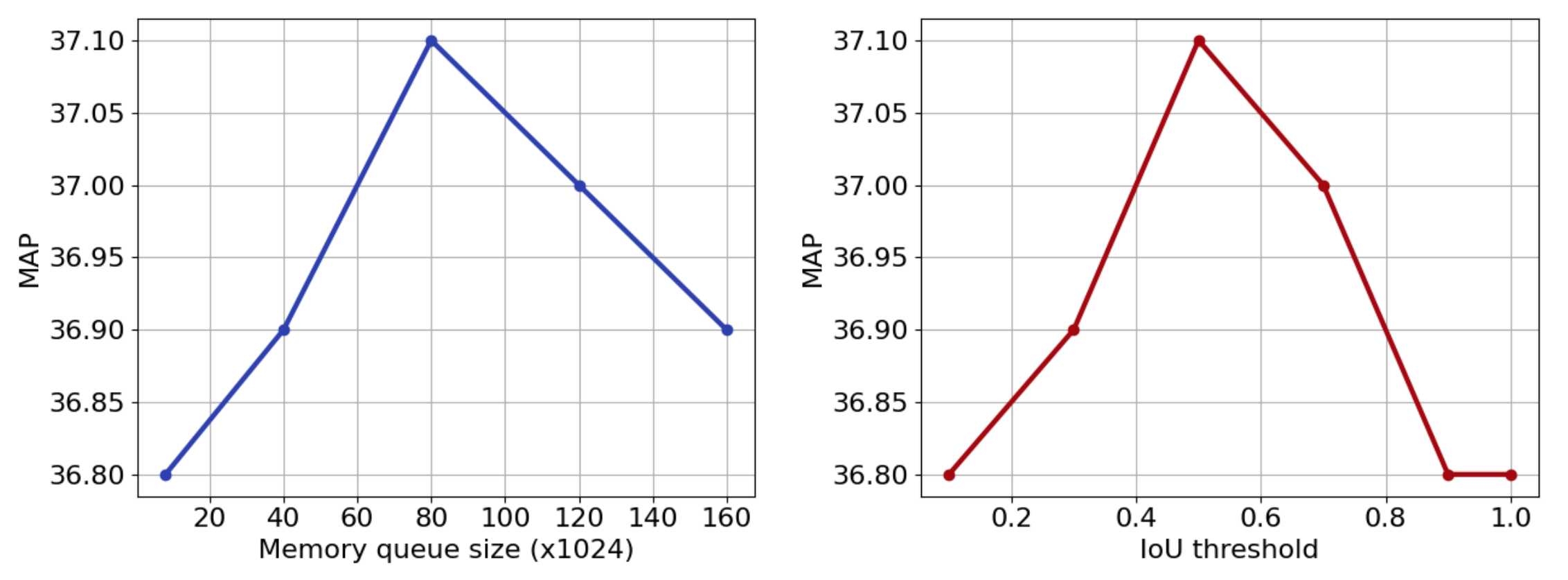}
\par\end{centering}
\caption{\label{fig:heterogeneous-1}Performance plots for different values
of memory queue size and IoU threshold. The optimal IoU threshold
is around 0.5, while the best memory queue size is $1024\times80$.}
\end{figure}

The performance curves plotted by varying the values for memory queue
size and IoU threshold are demonstrated in Figure \ref{fig:heterogeneous-1}.

\begin{table}
\begin{centering}
\begin{tabular}{c|c|c}
\hline 
Memory Size & Student & AP\tabularnewline
\hline 
1024{*}8 & \multirow{2}{*}{R18} & 36.8$^{+2.8}$\tabularnewline
1024{*}40 &  & 36.9$^{+2.9}$\tabularnewline
\textbf{1024{*}80} & FasterRCNN & \textbf{37.1$^{+3.1}$}\tabularnewline
1024{*}120 & \multirow{2}{*}{(34.0)} & 37.0$^{+3.0}$\tabularnewline
1024{*}160 &  & 36.9$^{+2.9}$\tabularnewline
\hline 
\end{tabular}
\par\end{centering}
\caption{\label{tab:banksize-1}Performance of Contrastive KD with different
memory sizes. The results show that the optimal memory size $K$ is
around 1024{*}80.}
\end{table}

\begin{table}
\begin{centering}
\begin{tabular}{c|c|c}
\hline 
IoU Threshold & Student & AP\tabularnewline
\hline 
0.1 & \multirow{2}{*}{R18} & 36.8$^{+2.8}$\tabularnewline
0.3 &  & 36.9$^{+2.9}$\tabularnewline
\textbf{0.5} & FasterRCNN & \textbf{37.1$^{+3.1}$}\tabularnewline
0.7 & \multirow{2}{*}{(34.0)} & 37.0$^{+3.0}$\tabularnewline
0.9 &  & 36.8$^{+2.8}$\tabularnewline
1.0 &  & 36.8$^{+2.8}$\tabularnewline
\hline 
\end{tabular}
\par\end{centering}
\caption{\label{tab:IoUthreshold}Performance of Contrastive KD with different
IoU thresholds for negative assignment. The results show that the
optimal IoU threshold is around 0.5.}
\end{table}

\subsection{Projection Head}

Recall that the critic function $g\left(r_{s},r_{t}\right)=\exp\left(\frac{f_{\theta}\left(r_{s}\right)\cdot f_{\theta}\left(r_{t}\right)}{\parallel f_{\theta}\left(r_{s}\right)\parallel\cdot\parallel f_{\theta}\left(r_{t}\right)\parallel}\cdot\frac{1}{\gamma}\right)$
utilizes a projection head $f_{\theta}$ to map the representations
to a lower dimension for both student and teacher. \cite{chen2020simple}
claimed that using a nonlinear projection head improves the representation
quality. However, this finding does not apply in our case. We observe
that linear projection head outperforms its nonlinear counterpart.
We assume this is because introducing nonlinearity into the projection
further complicates the learning process. The experiments are shown
in Table \ref{tab:projection-mapping}.

\begin{table}
\begin{centering}
\tabcolsep 0.02in{\scriptsize{}}%
\begin{tabular}{c|c|c}
\hline 
Method & Projection & AP\tabularnewline
\hline 
Baseline & N/A & 34.0\tabularnewline
\hline 
\multirow{2}{*}{CKD} & nonlinear & 36.7$^{+2.7}$\tabularnewline
 & \textbf{linear} & \textbf{37.1$^{+3.1}$}\tabularnewline
\hline 
\end{tabular}{\scriptsize\par}
\par\end{centering}
\caption{\label{tab:projection-mapping}Comparison between nonlinear and linear
projecting heads. Linear projection head outperforms its nonlinear
counterpart for our CKD.}
\end{table}

\subsection{Forming Contrastive Pairs for Heterogeneous Detectors}

As elaborated in the main paper, when dense prediction detector is
used as the student, the contrastive pairs are constructed using the
representations of the teacher's last fully connected layer and the
corresponding features from the last layer of student's classification
branch. However, the representations from student's localization branch
may also be used for CKD. We compare the performance of different
ways to construct contrastive pairs. The observation is that using
student's classification representations brings to the most gain.
We assume this is because the effectiveness of CKD is reflected mostly
on its classification ability. The results are shown in Table \ref{tab:contrastive-branch}.

\begin{table}
\begin{centering}
\begin{tabular}{c|c|c}
\hline 
Branch & Student & AP\tabularnewline
\hline 
\textbf{classification} & R18 & \textbf{35.8$^{+3.2}$}\tabularnewline
localization & RetinaNet & 34.3$^{+1.7}$\tabularnewline
combined & (32.6) & 35.1$^{+2.5}$\tabularnewline
\hline 
\end{tabular}
\par\end{centering}
\caption{\label{tab:contrastive-branch}Performance of Contrastive KD using
representations from different branches of the student. The results
show that using representation from student's classification branch
leads to the most performance gain. \textquotedblleft combined\textquotedblright{}
means summing up the corresponding representations from both heads
for forming contrastive pairs.}
\end{table}

\section{Localization Distillation with Uncertainty}

We show in Table \ref{tab:class-aware-regress} the superior performance
of our proposed class-aware localization distillation (elaborated
in Section 4.3.1 of the main paper) in contrast to the regular approach
which adopts L1 loss. As can be seen, our CAReg outperforms the regular
KD by a large margin.

\begin{table}
\begin{centering}
\tabcolsep 0.02in{\scriptsize{}}%
\begin{tabular}{c|c|c}
\hline 
Method & Class-aware & AP\tabularnewline
\hline 
Baseline & N/A & 34.0\tabularnewline
\hline 
\multirow{2}{*}{Regression} & N & 34.7$^{+0.7}$\tabularnewline
 & Y & \textbf{35.7$^{+1.7}$}\tabularnewline
\hline 
\end{tabular}{\scriptsize\par}
\par\end{centering}
\caption{\label{tab:class-aware-regress}Comparison between class-agnostic
and class-aware regression losses. 'N' means class-agnostic; 'Y' means
class-aware. Class-agnostic loss only distills the regression outputs
corresponding to the proposal's ground truth class, while class-aware
loss incorporates the uncertainty information by calculating the sum
of all regression outputs weighted by their corresponding class confidence.
The result shows a significant boost when applying our class-aware
loss.}
\end{table}

\section{Prediction Distillation for Heterogeneous Detectors}

Knowledge distillation using the prediction outputs for heterogeneous
detector pairs is not straightforward, since the loss functions used
during training are usually different, which causes the outputs to
carry different meanings. E.g., two-stage detectors use cross entropy
loss, and dense prediction detectors often adopt focal loss \cite{lin2018focal}.
We attempt to conduct KD on the prediction outputs of FasterRCNN teacher
and RetinaNet student by converting the student's outputs to make
it have the same meaning as the teacher's outputs. Specifically, we
apply softmax function on the class dimension of student logits, the
result is divided by its maximum on the class dimension. Then we extract
only the values for object classes to obtain class predictions, which
has the same dimension as the teacher's prediction outputs. The conversion
can be formulated by:

\[
\mathbf{P}_{s}=\frac{softmax\left(\mathbf{L}_{s}\right)}{max\left(softmax\left(\mathbf{L}_{s}\right)\right)}\left[1,...,C\right]
\]
where $\mathbf{L}_{s}\in R^{N\times C+1}$ is the logits from the
student detector, N is the batch size and C is the number of classes
(excluding background); $softmax$ is the softmax function performed
on the class dimension; max takes the maximum from the class dimension;
$\left[1,...,C\right]$ means take only the values for object classes.

The KD loss can be formulated as: $L_{cls}=-\frac{1}{N}\sum^{N}\mathbf{P}_{t}\log\mathbf{P}_{s}$,
where $\mathbf{P}_{s}\in R^{N\times C}$, $\mathbf{P}_{t}\in R^{N\times C}$
are the class scores of the student and the teacher, respectively.

\section{Generalization Ability of G-DetKD}

We conduct additional experiments to explore the generalization ability
of our G-DetKD for various detector architectures with different capacities.
The results in Table \ref{tab:change-student} shows our method consistently
improves the student's performances. Homogeneous detector pairs are
used.

\begin{table}
\begin{centering}
\tabcolsep 0.02in{\small{}}%
\begin{tabular}{c|c|c|c}
\hline 
{\small{}Model} & {\small{}Student} & {\small{}Teacher} & {\small{}AP}\tabularnewline
\hline 
\multirow{3}{*}{{\small{}Faster-RCNN-C4}} & {\small{}R18 (22.0)} & {\small{}R50 (34.8)} & {\small{}29.1$^{+7.1}$}\tabularnewline
 & {\small{}R50 (31.9)} & {\small{}R50 (34.8)} & {\small{}34.7$^{+2.8}$}\tabularnewline
 & {\small{}R101 (36.0)} & {\small{}R50 (34.8)} & {\small{}36.8$^{+0.8}$}\tabularnewline
\hline 
\multirow{3}{*}{{\small{}Faster-RCNN-Cascade}} & {\small{}R18 (36.5)} & {\small{}R50 (43.0)} & {\small{}40.4$^{+3.9}$}\tabularnewline
 & {\small{}R50 (40.3)} & {\small{}R50 (43.0)} & {\small{}42.5$^{+2.2}$}\tabularnewline
 & {\small{}R101 (42.5)} & {\small{}R50 (43.0)} & {\small{}43.3$^{+0.8}$}\tabularnewline
\hline 
\end{tabular}{\small\par}
\par\end{centering}
\caption{\label{tab:change-student} Our KD framework shows performance gains
for students with different structures and capacities. The values
in the parentheses indicate baseline APs.}
\end{table}

\section{Proofs}

In this section, we provide proofs for: (1) the optimal critic function
$g^{*}\text{\ensuremath{\left(r_{s},r_{t}\right)}}$ is proportional
to the ratio between the joint distribution $p\left(f_{\theta}\left(r_{s}\right),f_{\theta}\left(r_{t}\right)\right)$
and the product of marginal distributions $p\left(f_{\theta}\left(r_{s}\right)\right)p\left(f_{\theta}\left(r_{t}\right)\right)$.
i.e., $\text{\ensuremath{g^{*}\text{\ensuremath{\left(r_{s},r_{t}\right)}}}}\propto\frac{p\left(f_{\theta}\left(r_{s}\right),f_{\theta}\left(r_{t}\right)\right)}{p\left(f_{\theta}\left(r_{s}\right)\right)p\left(f_{\theta}\left(r_{t}\right)\right)}$;
(2) Minimizing our contrastive loss $L_{ckd}$ has the effect of maximizing
the lower bound on the mutual information (MI) between the teacher's
and student's latent representations. Our proof follows the standard
structure outlined in \cite{tian2020contrastive,oord2018representation}.

\subsection{Critic function}

Mutual information is defined as the $KL$ divergence between the
joint distribution and the product of marginal distribution of two
random variables:

\begin{align*}
MI\left(X;Y\right)= & D_{KL}\left(\mathrm{P}_{XY}\left(x,y\right)||\mathrm{P}_{X}\left(x\right)\mathrm{P}_{Y}\left(y\right)\right)\\
= & \sum_{\text{x,y}}\mathrm{P}_{XY}\left(x,y\right)\log\frac{\mathrm{P}_{XY}\left(x,y\right)}{\mathrm{P}_{X}\left(x\right)\mathrm{P}_{Y}\left(y\right)}\\
= & \mathbb{E}_{\mathrm{P}_{XY}}\log\frac{\mathrm{P}_{XY}\left(x,y\right)}{\mathrm{P}_{X}\left(x\right)\mathrm{P}_{Y}\left(y\right)}
\end{align*}
Thus, the first step of our proof is that the optimal critic function
$\text{\ensuremath{g^{*}\text{\ensuremath{\left(r_{s},r_{t}\right)}}}}$
is proportional to the ratio between the joint distribution and the
product of marginal distributions. Note that $g$ contains a learnable
projection mapping $f_{\theta}$, and here we denote $g^{*}$ as the
critic function with the optimal parameters $\theta^{*}$. We can
consider the distribution of positive pairs as $p_{pos}=p\left(r_{s},r_{t}\right)$
(the joint distribution) and the distribution of negative pairs as
$p_{neg}=p\left(r_{s}\right)p\left(r_{t}\right)$ (the product of
marginal distributions). Suppose the $\left\{ r_{s}^{i},r_{t}^{i}\right\} $
forms a positive sample pair (the representation of the same proposal
feature), all other teacher's representations $\left\{ r_{t}^{j}\right\} \left(i\neq j\right)$
form negative pairs with $\left\{ r_{s}^{i}\right\} $ (representations
of different proposal features). Namely, $\left\{ r_{s}^{i},r_{t}^{i}\right\} $
is a sample from $p_{pos}$ while all other pairs $\left\{ r_{s}^{i},r_{t}^{j}\right\} \left(i\neq j\right)$
are samples from $p_{neg}$. We denote the optimal probability to
be $p\left(pos=i\right)$. Thus, we can have the following equation:

\begin{align*}
p\left(pos=i\right)= & \frac{p_{pos}\left(r_{s}^{i},r_{t}^{i}\right)\prod_{n=0,n\neq i}^{k}p_{neg}\left(r_{s}^{i},r_{t}^{n}\right)}{\sum_{j=0}^{k}p_{pos}\left(r_{s}^{i},r_{t}^{j}\right)\prod_{n=0,n\neq j}^{k}p_{neg}\left(r_{s}^{i},r_{t}^{n}\right)}\\
= & \frac{p\left(r_{s}^{i},r_{t}^{i}\right)\prod_{n=1,n\neq i}^{k}p\left(r_{s}^{i}\right)p\left(r_{t}^{n}\right)}{\sum_{j=0}^{k}p_{pos}\left(r_{s}^{i},r_{t}^{j}\right)\prod_{n=0,n\neq j}^{k}p\left(r_{s}^{i}\right)p\left(r_{t}^{n}\right)}\\
= & \frac{\frac{p\left(r_{s}^{i},r_{t}^{i}\right)}{p\left(r_{s}^{i}\right)p\left(r_{t}^{i}\right)}}{\sum_{j=0}^{k}\frac{p\left(r_{s}^{i},r_{t}^{j}\right)}{p\left(r_{s}^{i}\right)p\left(r_{t}^{j}\right)}}\\
= & \frac{g^{*}\left(r_{s}^{i},r_{t}^{i}\right)}{\sum_{j=0}^{K}g^{*}\left(r_{s}^{i},r_{t}^{j}\right)}
\end{align*}
We first plug in $p_{pos}$ and $p_{neg}$, then divide the nominator
and denominator by $\prod_{n=1}^{k}p_{neg}\left(r_{s}^{i},r_{t}^{n}\right)$
at the same time, which leads to the final form of the equation. Note
that $g$ can be defined for either the original feature inputs $\left\{ r_{s},r_{t}\right\} $
or the latent representations $\left\{ f_{\theta}\left(r_{s}\right),f_{\theta}\left(r_{t}\right)\right\} $.
As the latent representations are used in practice, we will replace
$\left\{ r_{s},r_{t}\right\} $ by $\left\{ \gamma_{s},\gamma_{t}\right\} $
in following proofs (we denote $f_{\theta}\left(r\right)$ as $\gamma$
for simplicity). We can see that according to the definition of our
loss function, $\text{\ensuremath{g^{*}\text{\ensuremath{\left(r_{s},r_{t}\right)}}}}$
is actually proportional to $\frac{p\left(\gamma_{s},\gamma_{t}\right)}{p\left(\gamma_{s}\right)p\left(\gamma_{t}\right)}$.

\subsection{Maximizing the lower bound of MI}

As derived from above, $\text{\text{\ensuremath{g^{*}\text{\ensuremath{\left(r_{s},r_{t}\right)}}}}}\propto\frac{p\left(z_{s},z_{t}\right)}{p\left(z_{s}\right)p\left(z_{t}\right)}$,
we can then substitute the $\text{\ensuremath{g^{*}\text{\ensuremath{\left(r_{s},r_{t}\right)}}}}$
in our loss function by $\frac{p\left(z_{s},z_{t}\right)}{p\left(z_{s}\right)p\left(z_{t}\right)}$
, then we have the following expression:
\begin{align*}
L_{ckd}^{opt}= & -\mathbb{E}\log\left[\frac{g^{*}\left(r_{s}^{i},r_{t}^{i}\right)}{\sum_{j}^{K}g^{*}\left(r_{s}^{i},r_{t}^{j}\right)}\right]\\
= & -\mathbb{E}\log\left[\frac{\frac{p\left(\gamma_{s}^{i},\gamma_{t}^{i}\right)}{p\left(\gamma_{s}^{i}\right)p\left(\gamma_{t}^{i}\right)}}{\sum_{j}^{k}\frac{p\left(\gamma_{s}^{i},\gamma_{t}^{j}\right)}{p\left(\gamma_{s}^{i}\right)p\left(\gamma_{t}^{j}\right)}}\right]\\
= & \mathbb{E}\log\left[\frac{p\left(\gamma_{s}^{i}\right)p\left(\gamma_{t}^{i}\right)}{p\left(\gamma_{s}^{i},\gamma_{t}^{i}\right)}\sum_{j\neq i}^{K}\frac{p\left(\gamma_{s}^{i},\gamma_{t}^{j}\right)}{p\left(\gamma_{s}^{i}\right)p\left(\gamma_{t}^{j}\right)}+1\right]\\
\thickapprox & \mathbb{E}\log\left[\frac{p\left(\gamma_{s}^{i}\right)p\left(\gamma_{t}^{i}\right)}{p\left(\gamma_{s}^{i},\gamma_{t}^{i}\right)}K+1\right]\\
\geq & \log\left(K\right)-\mathbb{E}\log\left[\frac{p\left(\gamma_{s}^{i},\gamma_{t}^{i}\right)}{p\left(\gamma_{s}^{i}\right)p\left(\gamma_{t}^{i}\right)}\right]\\
= & \log\left(K\right)-\mathbb{E}_{p_{pos}}\log\left[\frac{p\left(\gamma_{s},\gamma_{t}\right)}{p\left(\gamma_{s}\right)p\left(\gamma_{t}\right)}\right]\\
= & \log\left(K\right)-MI\left(f_{\theta}\left(r_{s}\right);f_{\theta}\left(r_{t}\right)\right)
\end{align*}

As can be seen, minimizing $L_{ckd}$ can be interpreted as maximizing
the mutual information between $\left\{ z_{s},z_{t}\right\} $. We
can notice in the equation that larger $K$ leads to a tighter lower
bound, thus it is theoretically beneficial to set $K$ to be a very
large number. However, we experimentally find that it is not true
for our contrastive KD in object detection. The experiments are shown
in previous section.

\end{appendix}
\end{document}